\documentclass[10pt,twocolumn,letterpaper]{article}

\usepackage{iccv}              

\usepackage{graphicx}
\usepackage{cuted}
\usepackage{capt-of}
\usepackage{amsmath}
\usepackage{caption}
\usepackage{subcaption}
\usepackage{amssymb}
\usepackage{multibib}
\usepackage{wrapfig}
\usepackage{multirow}
\usepackage{multicol}
\usepackage{booktabs}
\usepackage{enumitem}
\usepackage{pifont}
\usepackage{xcolor}
\usepackage{lipsum}
\usepackage{float}

\newcommand{\ours}{HyperFlow}

\definecolor{iccvblue}{rgb}{0.21,0.49,0.74}
\usepackage[pagebackref,breaklinks,colorlinks,allcolors=iccvblue]{hyperref}

\def\paperID{11272} 
\def\confName{ICCV}
\def\confYear{2025}

\title{\ours{}: Gradient-Free Emulation of Few-Shot Fine-Tuning}

\author{
Donggyun Kim\qquad\qquad Chanwoo Kim\qquad\qquad Seunghoon Hong\vspace{0.1cm}\\
KAIST School of Computing\\
{\tt\small donggyun.kim@kaist.ac.kr, kcwkcw0802@kaist.ac.kr, seunghoon.hong@kaist.ac.kr}
}

\begin{document}
\maketitle
\begin{abstract}
While test-time fine-tuning is beneficial in few-shot learning, the need for multiple backpropagation steps can be prohibitively expensive in real-time or low-resource scenarios.
To address this limitation, we propose an approach that emulates gradient descent without computing gradients, enabling efficient test-time adaptation.
Specifically, we formulate gradient descent as an Euler discretization of an ordinary differential equation (ODE) and train an auxiliary network to predict the task-conditional drift using only the few-shot support set.
The adaptation then reduces to a simple numerical integration (\emph{e.g.,} via the Euler method), which requires only a few forward passes of the auxiliary network—no gradients or forward passes of the target model are needed.
In experiments on cross-domain few-shot classification using the Meta-Dataset and CDFSL benchmarks, our method significantly improves out-of-domain performance over the non-fine-tuned baseline while incurring only $6\%$ of the memory cost and $0.02\%$ of the computation time of standard fine-tuning, thus establishing a practical middle ground between direct transfer and fully fine-tuned approaches.
\end{abstract}    
\section{Introduction}
\label{sec:intro}

Few-shot classification aims to recognize novel classes of images using only a handful of labeled examples per class~\cite{fei2006one,lake2015human,vinyals2016matching,ravi2017optimization}.
Unlike conventional supervised learning approaches that use an extensive amount of labeled images for pre-defined tasks to train specialized models, few-shot learning approaches focus on adapting a pretrained model to novel tasks using a few examples, where the target classes and image domains can vary at test time.
This paradigm is vital for real-world scenarios where the image domain is distinctive and the labeled data is scarce or expensive to obtain, such as medical imaging~\cite{kotia2020few}, plant disease classification~\cite{argueso2020few}, and remote sensing~\cite{sun2021research}.
Even in the era of increasing general-purpose foundation models~\cite{bommasani2021opportunities}, few-shot learning is still beneficial to achieving optimal performance in domain-specific contexts~\cite{liu2024few,xu2024towards,madan2025revisiting}.

Recent advances in few-shot classification often rely on an additional \emph{fine-tuning} stage using the few examples given at test time.
Thanks to the simplicity and effectiveness of their design, embedding-based algorithms~\cite{vinyals2016matching,snell2017prototypical,bateni2020improved} have been predominant in few-shot classification tasks.
These methods classify images by learning a shared embedding space and assigning labels based on distances or similarities to reference points (\emph{e.g.,} class prototypes or exemplars) within that space.
While these approaches were originally designed to directly transfer a pretrained feature extractor to novel classes without adaptation, recent studies have consistently shown that additional fine-tuning at test time significantly improves performance, particularly when substantial domain discrepancies exist between training and testing scenarios~\cite{guo2020broader,phoo2021self,islam2021dynamic,oh2022understanding,hu2022pushing}.

However, despite its effectiveness, fine-tuning incurs substantial computational costs, both in memory and time, making it impractical for many real-world few-shot learning scenarios such as resource-constrained environments or real-time applications.
This high computational overhead primarily results from the repeated backpropagation required during gradient descent.
Therefore, achieving a balance between direct transfer and fine-tuning necessitates a computationally efficient adaptation mechanism that avoids gradient-based optimization entirely.
This perspective shares a high-level inspiration with parameter-generation methods~\cite{ha2017hypernetworks,knyazev2021parameter,peebles2022learning,schurholt2022hyper,schurhol2024ttowards}, which employ hypernetworks to generate model parameters.
However, most existing methods primarily focus on modeling parameter distributions for a pre-defined task with different architectures or training configurations, and relatively little attention has been paid to adapting pretrained parameters for novel classification tasks given only a few examples.

In this paper, we propose \emph{\ours{}}, a model-agnostic adaptation mechanism for few-shot classification that emulates fine-tuning without computing any gradients. 
To achieve this, we introduce an auxiliary \emph{drift} network that learns the parameter dynamics of the ordinary differential equations (ODEs) induced by the gradient descent (\emph{i.e.,} gradient flows).
As the gradient flow depends on the tasks, we amortize them by conditioning the drift network on a few support examples.
Since directly learning the ODEs in the full parameter space of modern neural networks is infeasible, we adopt a parameter-efficient fine-tuning (PEFT) approach~\cite{zaken2022bitfit}, which reduces the total parameters that need to be updated and thereby enables scalable training and inference.
In addition, we approximate the smooth gradient flows by interpolating the discrete gradient descent trajectories collected from an offline simulation on the meta-training dataset, enabling efficient training of the conditional drift network.
As a result, the fine-tuning stage can be replaced with a lightweight ODE-solver (\emph{e.g.,} Euler method) that involves only a few forward passes through the drift network.
The drift network can be significantly smaller and more computationally efficient than the target models, thus incurring negligible computational overhead compared to gradient-based fine-tuning methods.

We demonstrate our approach in cross-domain few-shot classification settings using widely adopted benchmarks such as Meta-Dataset~\cite{triantafillou2020meta} and CDFSL~\cite{guo2020broader}.
When applied to the state-of-the-art few-shot classification approach~\cite{hu2022pushing}, \ours{} can significantly improve the out-of-domain generalization by only using $6\%$ of the memory cost and $0.02\%$ of the computation time compared to conventional fine-tuning.
As a result, \ours{} offers a range of trade-offs between the computation cost and the performance and thus suggests the practical middle-ground between the direct transfer and fine-tuning approaches.
\section{Preliminary}
\label{sec:preliminary}

\paragraph{Problem Setup}
Few-shot classification (FSC) aims to recognize novel classes of images using only a few labeled examples per class.
To achieve this, a model is first trained on a base dataset $\mathcal{D}_\text{base} = \{(x_i, y_i)\}_{i=1}^{N_\text{base}}$, often called as ``meta-training" dataset, which consists of many labeled examples for known base classes $\mathcal{C}_\text{base}$.
During testing, the model is evaluated on multiple classification tasks involving novel classes $\mathcal{C}_\text{novel}$, where $\mathcal{C}_\text{base} \cap \mathcal{C}_\text{novel} = \phi$.
To learn the novel classes of each test task $\mathcal{T}$, the model receives a small set of labeled examples $\mathcal{S}_\mathcal{T} = \{(x_j, y_j)\}_{j=1}^{N_\mathcal{T}}$ called the \emph{support set}.
Then, the model is tested to classify unlabeled images within the same task, called the \emph{query set}.
In conventional $N$-way $K$-shot setting~\cite{vinyals2016matching,finn2017model}, each task consists of $N$ classes, and $K$ labeled images per class are given as the support set.

Additionally, we consider the more challenging \emph{cross-domain} few-shot classification problem, where the testing images originate from domains different from those of the base dataset $\mathcal{D}_\text{base}$ (\emph{e.g.,} domain shifts from natural images to X-ray or satellite imagery).
In this setup, the model must equip strong generalization ability and flexible adaptation mechanism to address both novel classes and out-of-domain images.

\paragraph{P$>$M$>$F Pipeline for FSC}
While various approaches have been proposed to tackle the FSC problem, a common training pipeline involves one or more of three distinct stages—Pre-training, Meta-training, and Fine-tuning—collectively known as P$>$M$>$F~\cite{hu2022pushing}. In the pre-training stage, general-purpose representations are learned using backbones trained with self-supervised methods on large-scale datasets (\emph{e.g.,} the DINO backbone~\cite{caron2021emerging} pre-trained on ImageNet~\cite{deng2009imagenet}).
During the meta-training stage, the model acquires prior knowledge about the underlying task distribution, typically by simulating few-shot learning episodes using the base dataset $\mathcal{D}_\text{base}$~\cite{vinyals2016matching}.
Finally, the fine-tuning stage further adapts the model to test tasks using the provided support set, enabling flexible adaptation to novel classes and unseen domains.
During the fine-tuning stage, parameter-efficient fine-tuning (PEFT) techniques~\cite{zaken2022bitfit,hu2022lora} can be applied to reduce over-fitting to the support set.

\section{Method}
\label{sec:method}
\begin{figure*}
    \centering
    \includegraphics[width=\linewidth]{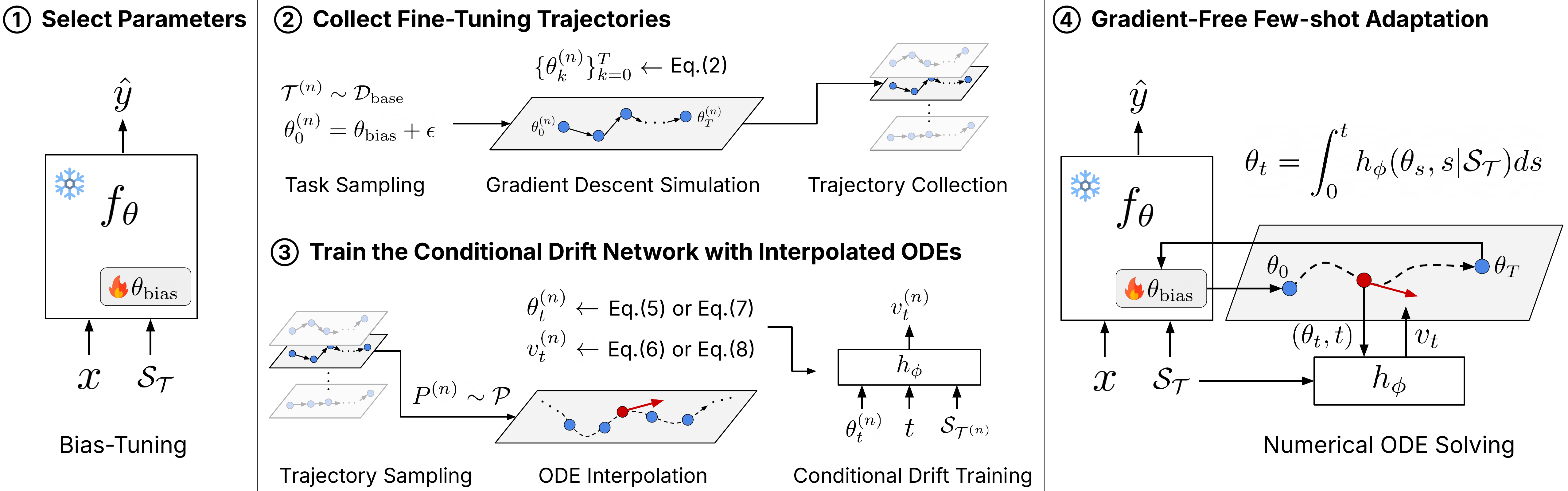}
    \caption{
    An overview of \ours{}.
    (1) For scalable computation in the parameter space, we select the bias parameters of the target model $f_\theta$ to be updated.
    (2) We collect $T$-step fine-tuning trajectories by simulating gradient descent on the bias parameters, using the base dataset where the target model has been trained.
    (3) After collecting the trajectories, we smoothly interpolate them using either linear or piecewise-cubic flow objective, then train the conditional drift network $h_\phi$ on the continuous time interval $[0, T]$.
    (4) At test time, we employ a numerical ODE solver (\emph{e.g.,} Euler method) with a few forward passes of $h_\phi$ to adapt the bias parameters on the support set.
    }
    \label{fig:main_figure}
    \vspace{-0.2cm}
\end{figure*}

In this work, we focus on the \emph{fine-tuning} stage of the training pipeline of few-shot classification (FSC).
While it has been consistently reported that fine-tuning is beneficial for FSC~\cite{guo2020broader,phoo2021self,islam2021dynamic,oh2022understanding,hu2022pushing}, especially in cross-domain scenarios, it necessitates the computation of expensive backpropagation throughout the network.
This can be problematic in practical applications of FSC where rapid adaptation is required or only low resources are allowed for adaptation.
To accelerate the few-shot adaptation of a model, we propose \ours{}, which replaces the fine-tuning stage with a much cheaper process.
See figure~\ref{fig:main_figure} for an overview of our method.

\subsection{\ours{}: Emulating Gradient Flows}
\label{sec:motivation}
Fine-tuning involves gradient descent on the model parameters.
Let $f_\theta$ be the model with parameters $\theta$ and $\mathcal{L}$ be a loss function (\emph{e.g.,} cross-entropy).
Given a test task $\mathcal{T}$ with a corresponding support set $\mathcal{S}_\mathcal{T} = \{(x_i, y_i)\}_{i=1}^{N_\mathcal{T}}$, the objective of few-shot adaptation is to minimize the loss over the support set.
\begin{equation}
L_\mathcal{T}(\theta)= \sum_{(x_i, y_i) \in \mathcal{S}_\mathcal{T}} \mathcal{L}\left(f_\theta(x_i), y_i\right).
\label{eqn:loss_objective}
\end{equation}
This is achieved by performing gradient descent that iteratively updates $\theta$ from its initialization $\theta_0 = \theta_\text{init}$ as follows:
\begin{equation}
\theta_k \leftarrow \theta_{k-1} - \lambda \cdot g\left(\nabla_\theta L_\mathcal{T}(\theta_{k-1})\right), \quad k=1, 2, \cdots, T,
\label{eqn:gradient_descent}
\end{equation}
where $\theta_k$ denotes the parameters at $k$-th iteration, $\lambda > 0$ denotes the learning rate, and $g$ is an optimizer (\emph{e.g.,} Adam~\cite{kingma2015adam}).
However, in resource-constrained environments, computing the gradient $\nabla_\theta L_\mathcal{T}(\theta)$ can be burdensome, especially when the network $f_\theta$ is deep and large as foundation models.

Our key idea is to emulate the gradient descent updates (Eq.~\eqref{eqn:gradient_descent}) with a \emph{gradient-free} ODE-solving process.
As already remarked by prior studies~\cite{elkabetz2021continuous,bu2021dynamical,merkulov2020stochastic}, the gradient descent updates can be seen as an Euler discretization of the corresponding ordinary differential equation (ODE), which is also known as the gradient flow~\cite{santambrogio2017euclidean}. For example, the gradient flow of task $\mathcal{T}$ with the simplest gradient descent optimizer (\emph{i.e.,} $g(\theta) = \theta$) can be expressed as follows:
\begin{equation}
\frac{d\theta(t)}{dt} = -\frac{\partial L_\mathcal{T}(\theta(t))}{\partial \theta}.
\label{eqn:gradient_flow}
\end{equation}
Inspired by Neural ODEs~\cite{chen2018neural}, we approximate the ODE function using a drift network $h_\phi$.
Then, we can emulate the gradient descent by a numerical integration method (\emph{e.g.,} the Euler method), which only requires several forward steps of $h_\phi$.

Since the gradient flow differs by tasks, we employ a \emph{conditional} drift network $h_\phi(\theta, t | \mathcal{S}_\mathcal{T})$ to predict the task-specific drift $\frac{d\theta(t)}{dt}$ of task $\mathcal{T}$ conditioned on the support set $\mathcal{S}_\mathcal{T}$.
To train the conditional drift network, we perform an extra training stage using the base dataset as follows:
\begin{equation}
\min_\phi~\mathbb{E}_{t, \mathcal{T}} \left[ \left\|h_\phi(\theta_t, t| \mathcal{S}_\mathcal{T}) - \left(-\frac{\partial L_\mathcal{T}(\theta(t))}{\partial \theta}\right) \right\|^2 \right],
\label{eqn:ode_objective}
\end{equation}
where the timestep $t$ is uniformly sampled from $[0, T]$ and the training task $\mathcal{T}$ is sampled from the base dataset $\mathcal{D}_\text{base}$.
As the base dataset is often assumed to be diverse enough to cover the underlying task distribution, the trained drift network $h_\phi$ can generalize to novel tasks.
In Section~\ref{sec:ablation_study}, we study how the diversity of the base dataset affects the generalization ability of $h_\phi$.

\begin{figure*}[!t]
    \centering
    \includegraphics[width=\linewidth]{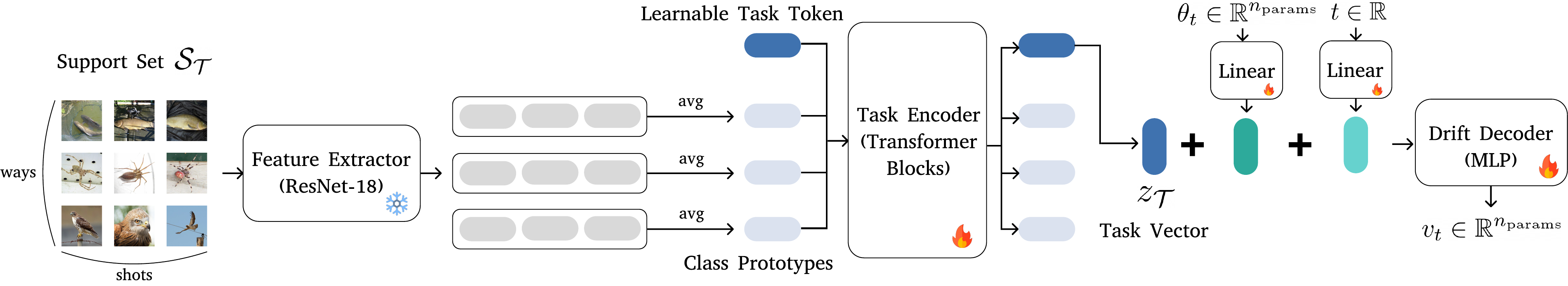}
    \caption{The architecture of the conditional drift network $h_\phi$ of \ours{}.}
    \label{fig:drift_architecture}
    \vspace{-0.2cm}
\end{figure*}

\subsection{Target Parameters Selection}
\label{sec:target_parameters_selection}
A straightforward challenge in learning the gradient flow with a drift network is that the number of target network parameters is often too large.
For example, vision transformers~\cite{dosovitskiy2021image}, a widely used neural network architecture in computer vision, typically have millions to billions of parameters, making it nearly infeasible to directly treat them as inputs and outputs of another neural network.
However, recent studies on parameter-efficient fine-tuning (PEFT) show that updating a tiny set of parameters is sufficient to modulate large pre-trained backbones~\cite{rebuffi2017learning,zaken2022bitfit,hu2022lora}.
Therefore, we reduce the number of parameters considered in the gradient flows by selecting a tiny subset of the whole parameters $\theta$.

Specifically, we adopt bias-tuning~\cite{zaken2022bitfit} that updates only the bias parameters of the model $f_\theta$.
In the few-shot learning context, bias-tuning has shown to be extremely parameter-efficient and robust to over-fitting for adapting large transformers to unseen vision tasks~\cite{kim2023universal,kim2024chameleon,chen2023vision}.
Also, it adds no extra computation overhead to the original model during inference.
By further selecting a subset of bias parameters (\emph{e.g.,} those from qkv-projection layers of the attention blocks), we can reduce the number of target parameters to be sufficiently tangible for the drift network (\emph{e.g.,} thousands) without losing much adaptability at the few-shot adaptation phase.

Once we have selected the feasible number of target parameters, another benefit of emulating gradient descent with the conditional drift network is that the computation involved in few-shot adaptation no longer depends on the architecture of the original network $f_\theta$.
For example, in conventional fine-tuning, even if we update a small subset of parameters, such as bias parameters, we still need to backpropagate the network until the layers have the parameters.
However, the drift network $h_\phi$ can be designed to produce an arbitrary set of parameters at once with a single forward operation.
This makes our approach benefit from PEFT techniques in reducing the computation cost, \emph{i.e.,} the parameter-efficiency directly translates to the computation-efficiency.
In Section~\ref{sec:cost_analysis}, we provide a detailed analysis of the computation efficiency of \ours{} compared to conventional fine-tuning.

\subsection{Drift Training with Smooth ODEs}
\label{sec:objective_odes}
Training the conditional drift network using the continuous-time ODE objective (Eq.~\eqref{eqn:ode_objective}) requires expensive computation of the gradient $\nabla_\theta L_\mathcal{T}(\theta)$ at arbitrary timestep $t \in [0, T]$ as well as the simulation of the gradient flow until $t$.
To make the training process more efficient, we first collect a fixed number of fine-tuning trajectories by simulating the gradient descent on the base dataset.
Specifically, for each task $\mathcal{T}^{(n)}$ constructed from the base dataset $\mathcal{D}_\text{base}$, a discretized trajectory $P^{(n)} = \{\theta_0^{(n)}, \theta_1^{(n)}, \cdots, \theta_T^{(n)}\}$ is generated by applying $T$ gradient descent steps (Eq.~\eqref{eqn:gradient_descent}) on the initial parameters $\theta_\text{init}^{(n)}$.
We repeat this process multiple times by randomly perturbing the initialization $\theta_\text{init}^{(n)}$, yielding a dataset of trajectories $\mathcal{P} = \{P^{(n)}\}_{n=1}^{N_\text{traj}}$.
During training, we approximate the continuous gradient flows by smoothly interpolating the discretized trajectories to compute the intermediate parameter $\theta_t$ and its drift $v_t$ at arbitrary timestep $t$.
We propose two objective ODEs that interpolate the discretized gradient flows.
Below, we omit the superscript of $\theta^{(n)}$ for notational simplicity.

\vspace{-0.1cm}
\paragraph{Linear Flow (\ours{}-L)}
A linear flow is generated by linearly interpolating the initial and final points of each trajectory~\cite{lipman2023flow,liu2023flow}.
Specifically, given the two endpoints $\theta_0$ and $\theta_T$ of a trajectory, the point $\theta_t$ and its drift $v_t$ (time-derivative) at timestep $t \in [0, T]$ is computed as follows:
\begin{align}
    \theta_t &= (1 - t/T) \cdot \theta_0 + t/T \cdot \theta_T, \\
    v_t &= \theta_T - \theta_0.
    \label{eqn:linear_flow}
\end{align}
Note that this objective does not use any intermediate points $\{\theta_k\}_{k=1}^{T-1}$.
As it promotes the straightness of the learned flow, it encourages the drift network to solve the ODE with only a few Euler steps.
However, the straight path emulated by Eq.~\eqref{eqn:linear_flow} may deviate from the true trajectory that follows the loss surface, so the flexibility of \ours{} can be limited by the oversimplified ODEs.

\vspace{-0.1cm}
\paragraph{Piecewise-Cubic Flow (\ours{}-C)}
To cope with more complex loss surfaces, we also propose a non-linear flow objective that exploits the full trajectories.
A straightforward way to extend the linear flow with intermediate points is to exactly follow the simulated gradient descent updates, \emph{i.e.,} define piecewise-linear flows.
However, such discrete transitions in the velocity field are difficult to model by the conditional drift network and can cause over-fitting to training trajectories.
Thus, we create smooth ODEs that pass through the discrete points using cubic splines.

To compute the interpolated point and its drift, the piecewise-cubic flow defines local cubic curves passing four neighboring points at each timestep.
Specifically, we divide the time interval $[0, T]$ into $T$ segments $[0, 1], [1, 2], \cdots, [T-1, T]$, then interpolate each segment $[k-1, k]$ with a cubic Hermite spline~\cite{hermite1863remarque,de1978practical} that requires the two endpoints $\theta_{k-1}, \theta_{k}$ and their tangents $m_{k-1}, m_{k}$ to compute the coefficients of the cubic curve.
Since we uniformly divided the time interval, we can employ the Catmull-Rom spline~\cite{catmull1974class} that sets the tangents using two additional points $\theta_{k-2}, \theta_{k+1}$.
Then, the point $p_t$ and its drift $v_t$ in the segment $[k-1, k]$ can be computed as follows:
\begin{align}
    \theta_t &= a_k t^3 + b_k t^2 + c_k t + d_k, \\
    v_t &= 3 a_k t^2 + 2 b_k t + c_k,
\end{align}
where the coefficients $a_k, b_k, c_k, d_k$ are computed by the four points $\theta_{k-2}, \theta_{k-1}, \theta_{k}, \theta_{k+1}$ as follows:
\begin{align}
    a_k &= -\frac{1}{2}\theta_{k-2} + \frac{3}{2}\theta_{k-1} - \frac{3}{2}\theta_{k} + \frac{1}{2}\theta_{k+1} \\
    b_k &= \theta_{k-2} -\frac{5}{2}\theta_i + 2\theta_{k} - \frac{1}{2}\theta_{k+1} \\
    c_k &= -\frac{1}{2}\theta_{k-2} + \frac{1}{2}\theta_{k} \\
    d_k &= \theta_{k-1}
\end{align}
For interpolating the first and the last segment, we set the sentinels as $\theta_{-1} = \theta_0$ and $\theta_{T+1} = \theta_T$.

\subsection{Conditional Drift Network Architecture}
\vspace{-0.1cm}
The conditional drift network $h_\phi$ consists of three submodules: feature extractor, task encoder, and the drift decoder (see Figure~\ref{fig:drift_architecture} for an illustration).
Since the target drift $v_t$ of the ODE depends on the task $\mathcal{T}$, we need an effective task encoding mechanism.
To this end, we first create class prototypes by encoding the support images using a feature extractor and aggregating them for each class.
For the feature extractor, we use a ResNet-18~\cite{he2016deep} pretrained on ImageNet-1k and freeze it during training.
Then, a task encoder further aggregates the class prototypes to produce a single task vector $z_\mathcal{T}$.
Since the number of classes (ways) can vary by tasks, we implement the task encoder with a 2-layer transformer~\cite{vaswani2017attention} that can process variable length of inputs.
Specifically, we introduce a learnable task token and encode it with the class prototypes by the task encoder, then use its final output embedding as a task vector.
Finally, we add the task vector $z_\mathcal{T}$ with the linear embeddings of the vectorized target parameters $\theta_t$ and timestep $t$, then pass it to the 4-layer MLP drift decoder to produce the drift prediction $v_t$.
\vspace{-0.1cm}
\vspace{-0.3cm}
\section{Related Work}
\label{sec:related_work}
\vspace{-0.1cm}
\paragraph{Metric-Based Few-Shot Learning}
Few-shot learning aims to adapt models to handle new tasks with minimal labeled data.
A predominant strategy of \emph{few-shot classification}, where each task is defined by a set of image classes~\citep{fei2006one, lake2015human}, is the metric-based learning that assigns the classes based on the distance on an embedding space learned by a feature extractor~\citep{vinyals2016matching,snell2017prototypical}.
While early methods of this type typically transfer the feature extractor to novel classes without additional adaptation, several studies~\citep{guo2020broader, phoo2021self, islam2021dynamic, oh2022understanding, hu2022pushing} have shown that test-time fine-tuning offers substantial improvements in cross-domain scenarios.
In particular, \citet{hu2022pushing} introduced P$>$M$>$F, a practical three-stage pipeline—self-supervised pretraining, episodic meta-learning, and test-time fine-tuning—that significantly boosts performance on out-of-domain tasks.

\vspace{-0.4cm}
\paragraph{Meta-Optimizers and Gradient-Based Meta-Learning.}
Another branch of few-shot learning research focuses on training meta-optimizers or finding good parameter initializations to quickly adapt to new tasks \citep{andrychowicz2016learning, finn2017model, ravi2017optimization}.
Although these methods eliminate hand-crafted optimizers and encourage rapid adaptation, they require gradients for adaptation, increasing computational overhead and hyperparameter sensitivity. 
Recently, \citet{zhang2024metadiff} proposed MetaDiff as a gradient-free approach that models the adaptation process via a denoising diffusion model~\cite{ho2020denoising}.
Similarly, \citet{du2023protodiff} and \citet{zhang2022metanode} focus on the prototypical network architecture and meta-learn the prototype refinement process using a diffusion model~\cite{ho2020denoising} and a Neural ODE~\cite{chen2018neural}, respectively.
By contrast, our method proposes a model-agnostic adaptation mechanism by emulating the gradient flows on the parameter space of the target model.

\vspace{-0.3cm}
\paragraph{Hypernetworks and Parameter Generation}
Parameter generation methods aim to generate model parameters using another neural network, often called a ``hypernetwork"~\citep{ha2017hypernetworks}.
These methods primarily aim to generate neural network parameters that generalize across various architectures or training configurations rather than adapting to novel tasks with a few examples~\cite{peebles2022learning,schurhol2024ttowards,knyazev2021parameter,schurholt2022hyper}.
Recently, \citet{jin2024conditional} and \citet{liang2024make} have proposed few-shot parameter generation methods for natural language processing (and style transfer) and policy learning, respectively.
Both rely on latent diffusion models~\cite{rombach2022high} operating in a parameter space learned via autoencoders.
Our approach, by contrast, explicitly learns the gradient flows in the original parameter space, which avoids the need for a latent autoencoder.

\section{Experiments}
\label{sec:experiments}
\vspace{-0.1cm}

The main advantage of \ours{} over existing few-shot learning approaches is that it strikes a middle ground between direct-transfer (without fine-tuning) approaches and fine-tuning-based approaches, offering both improved performance and reduced computational cost to each.
To demonstrate its robustness, we evaluate \ours{} under the cross-domain few-shot classification scenario.
We compare \ours{} with P$>$M$>$F~\citep{hu2022pushing}, one of the state-of-the-art methods that includes a few-shot fine-tuning stage.

\begin{table*}[!t]
\caption{Few-shot classification results on Meta-Dataset benchmark. As a gradient-free adaptation mechanism, \ours{} strikes the middle ground between directly transferring the ProtoNet and fine-tuning it during test time.}
\label{tab:pmf_md_table}
\vspace{-0.4cm}
\begin{center}
    \renewcommand{\arraystretch}{1.2}
    \renewcommand{\aboverulesep}{0pt}
    \renewcommand{\belowrulesep}{0pt}
    \setlength\tabcolsep{4.5pt}
    \footnotesize
    \begin{tabular}{c|ccc|cc|cccccccc}
        \toprule
        \multirow{2}{*}{Model Variant} &
        \multicolumn{3}{c|}{Average} &
        \multicolumn{2}{c|}{Out-of-Domain} &
        \multicolumn{8}{c}{In-Domain}
         \\
        \cmidrule{2-14}
        &
        Total & OOD & ID &
        Sign & COCO &
        Acraft & CUB & DTD & Fungi &
        Flower & Inet & Omglot & QDraw
        \\
        \midrule

        ProtoNet &
	77.22 & 54.38 & 82.93 & 
	54.02 & 54.74 & 
	86.93 & 90.62 & 80.58 & 73.26 & 
	91.81 & 72.21 & 89.87 & 78.13 \\

        \midrule
        
        ProtoNet + \ours{}-L \textbf{(Ours)} &
	79.10 & 58.97 & 84.13 & 
	59.81 & 58.13 & 
	86.93 & 91.03 & 84.61 & 73.30 & 
	95.86 & 73.03 & 89.96 & 78.34 \\
        
        ProtoNet + \ours{}-C \textbf{(Ours)} &
	79.26 & 61.09 & 83.88 & 
	65.29 & 56.88 & 
	86.24 & 90.77 & 83.96 & 73.34 & 
	95.77 & 73.07 & 89.71 & 78.16 \\

        \midrule
        
        ProtoNet + Bias-Tuning &
	82.43 & 74.97 & 84.29 & 
	90.30 & 59.65 & 
	87.60 & 91.05 & 85.33 & 72.66 & 
	95.32 & 73.76 & 90.17 & 78.43 \\
        
        ProtoNet + Full-Tuning &
	81.96 & 74.48 & 83.84 & 
	90.97 & 57.99 & 
	89.58 & 90.53 & 85.79 & 68.79 & 
	95.49 & 72.01 & 91.85 & 76.64 \\

        \bottomrule
    \end{tabular}
\end{center}
\vspace{-0.3cm}
\end{table*}
\begin{table*}[!t]
\caption{Few-shot classification results on CDFSL benchmark. \ours{} generalizes well on the out-of-domain tasks with various shots.}
\label{tab:pmf_cd_table}
\vspace{-0.4cm}
\begin{center}
    \renewcommand{\arraystretch}{1.2}
    \renewcommand{\aboverulesep}{0pt}
    \renewcommand{\belowrulesep}{0pt}
    \setlength\tabcolsep{2.8pt}
    \footnotesize
    \begin{tabular}{c|ccc|ccc|ccc|ccc|ccc}
        \toprule
        \multirow{2}{*}{Model Variant} &
        \multicolumn{3}{c|}{Average} &
        \multicolumn{3}{c|}{EuroSAT} &
        \multicolumn{3}{c|}{CropDisease} &
        \multicolumn{3}{c|}{ISIC} &
        \multicolumn{3}{c}{ChestX} \\
        \cmidrule{2-16}

        & 
        5w5s & 5w20s & 5w50s &
        5w5s & 5w20s & 5w50s &
        5w5s & 5w20s & 5w50s &
        5w5s & 5w20s & 5w50s &
        5w5s & 5w20s & 5w50s \\
        
        \midrule

        ProtoNet &
	56.71 & 62.14 & 64.14 & 
	81.02 & 86.76 & 87.95 & 
	85.67 & 91.14 & 92.14 & 
	35.27 & 42.37 & 45.84 & 
	24.90 & 28.29 & 30.61 \\

        \midrule
        
        ProtoNet + \ours{}-L \textbf{(Ours)} &
	61.14 & 66.34 & 68.10 & 
	85.58 & 90.63 & 91.70 & 
	91.80 & 94.90 & 95.33 & 
	41.75 & 50.24 & 53.18 & 
	25.45 & 29.57 & 32.18 \\
        
        ProtoNet + \ours{}-C \textbf{(Ours)} &
	61.76 & 67.22 & 69.01 & 
	86.35 & 90.43 & 92.14 & 
	92.56 & 95.37 & 95.83 & 
	43.83 & 53.40 & 56.32 & 
	25.31 & 29.67 & 31.98 \\
    
        \midrule
        
        ProtoNet + Bias-Tuning &
	64.07 & 72.37 & 74.51 & 
	85.42 & 93.30 & 95.45 & 
	92.60 & 97.71 & 98.80 & 
	51.87 & 64.47 & 67.34 & 
	26.40 & 33.99 & 36.46 \\
        
        ProtoNet + Full-Tuning &
	64.69 & 71.48 & 76.58 & 
	85.43 & 91.90 & 96.81 & 
	92.62 & 97.55 & 99.22 & 
	53.02 & 61.05 & 69.48 & 
	27.69 & 35.43 & 40.82 \\

        \bottomrule
    \end{tabular}
\end{center}
\vspace{-0.3cm}
\end{table*}


\vspace{-0.1cm}
\subsection{Experimental Setup}

\vspace{-0.1cm}
\paragraph{Datasets}
Following the training and evaluation protocol of \citet{hu2022pushing}, we use the Meta-Dataset benchmark \citep{triantafillou2020meta}, which consists of ten public datasets spanning diverse domains: ImageNet-1k, Omniglot, FGVC-Aircraft, CUB-200-2011, Describable Textures, QuickDraw, FGVCx Fungi, VGG Flower, Traffic Signs, and MSCOCO.
We adopt the 8 in-domain setup, treating the training splits of the first eight datasets as the \emph{base} dataset and evaluate on the test splits of all ten; notably, Traffic Signs and MSCOCO serve as out-of-domain tasks.
Each test task in Meta-Dataset follows a \emph{various-way various-shot} protocol, where both the number of classes and the number of support images per class are randomly chosen.
In addition, we include four out-of-domain datasets from the CDFSL benchmark \citep{guo2020broader}: CropDisease, EuroSAT, ISIC, and ChestX. Here, tasks are constructed in 5-way with 5-, 20-, and 50-shots.

\vspace{-0.3cm}
\paragraph{Baselines}
We choose the \textbf{ProtoNet}~\citep{snell2017prototypical} variant proposed by \citet{hu2022pushing} as our target model.
It is first pretrained on ImageNet-1k using DINO~\citep{caron2021emerging} objective, followed by meta-training on the eight base datasets through the prototypical network framework~\citep{snell2017prototypical}.
This model is directly transferred to novel tasks without test-time adaptation.
On the other hand, \textbf{ProtoNet + Full-Tuning} involves test-time fine-tuning stage, wherein it generates a pseudo-query set by augmenting the support set and fine-tunes the model to classify these pseudo-queries using the original support set.
We fine-tune the ProtoNet with 50 gradient descent steps, with searching the learning rate following \citet{hu2022pushing}.
Since these baselines represent two computational extremes, we propose to apply \ours{} to adapt the ProtoNet rather than fine-tuning.
We report both variants with linear (\textbf{ProtoNet + \ours{}-L}) and piecewise-cubic (\textbf{ProtoNet + \ours{}-C}) ODE objectives.
Because \ours{} only updates the bias parameters of the backbone, we also include the \textbf{ProtoNet + Bias-Tuning}, a variant of P$>$M$>$F that only updates the bias parameters, whose performance serves as an upper bound of ours.
For all models, we use the ViT-small~\citep{dosovitskiy2021image} backbone for the feature extractor of the prototypical network.
For \ours{} and the Bias-Tuning variant, we update the bias parameters of qkv-projection from each attention layer of the backbone transformer, which totals 13,824 parameters.

\vspace{-0.4cm}
\paragraph{Implementation Details}
We train the conditional drift network by collecting fine-tuning trajectories from the ProtoNet model on the eight training domains of the Meta-Dataset.
Specifically, we sample 500 episodes per domain and perform 50 gradient descent steps with the Adam optimizer~\citep{kingma2015adam}.
For each episode, we apply ten different random perturbations $\epsilon \sim \mathcal{N}(0, 0.04)$ to the initial parameters, yielding a total of 40,000 trajectories (each of length 51).
We also collect 80 additional trajectories from the validation split for early stopping.
When training the linear flow variant, we use only the initial and final parameters of each trajectory to interpolate the ODE.
For the cubic flow variant, we leverage all 51 parameter points in each trajectory.
During evaluation, we apply the Euler method with 50 steps and treat the step size as a hyper-parameter, tuning it in the same manner that P$>$M$>$F tunes its learning rate.

\begin{figure*}
    \centering
    \includegraphics[width=0.95\linewidth]{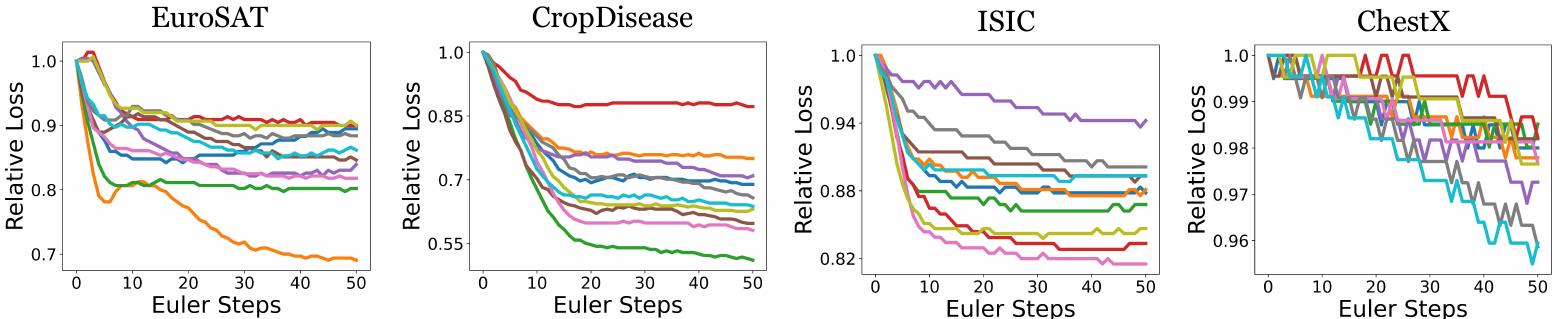}
    \caption{Intermediate losses on CDFSL tasks (ten tasks for each domain) during the parameter adaptation of \ours{}-C. Curves with different colors indicate different tasks within each domain, the x-axis corresponds to the number of steps taken in the Euler solver, and the y-axis corresponds to the relative loss normalized by the initial value.
    This shows that the conditional drift network of \ours{} generalizes well to the out-of-domain classification tasks by consistently moving the parameters to the regions with lower loss values.
    }
    \label{fig:loss_trajectory}
    \vspace{-0.2cm}
\end{figure*}

\vspace{-0.1cm}
\subsection{Main Result}
Table~\ref{tab:pmf_md_table} and Table~\ref{tab:pmf_cd_table} present our main results on the Meta-Dataset and CDFSL benchmarks, respectively.
On average, the performance of ProtoNet + \ours{} stands between the vanilla ProtoNet and the fine-tuned ProtoNet variants, as expected.
Remarkably, we find that incorporating either the linear or piecewise-cubic variant of \ours{} into the vanilla ProtoNet model substantially boosts out-of-domain performance.
This improvement indicates that the conditional drift network is not merely memorizing the fine-tuning trajectories but is instead learning a shared geometry of the underlying loss surfaces across tasks.
To further investigate out-of-domain generalization, Figure~\ref{fig:loss_trajectory} shows the intermediate losses on out-of-domain tasks during the ODE solving process of \ours{}-C.
The consistently decreasing loss curves indicate that the trained conditional drift network effectively generalizes to out-of-domain tasks, successfully driving the parameters toward lower-loss regions.

However, for in-domain tasks, the benefits of applying \ours{} to the vanilla ProtoNet are minimal, with notable improvements only for DTD, Flower, and ImageNet.
This is likely due to the high transferability of the vanilla ProtoNet backbone to in-domain tasks, where fine-tuning variants also show limited additional gains, consistent with the findings in \citet{hu2022pushing}.
In fact, ProtoNet + \ours{} performs similarly to the fine-tuning variants on most in-domain tasks, suggesting that the small improvements of \ours{} in these cases are more likely due to the inherent limitations of fine-tuning rather than a lack of adaptation flexibility of \ours{}.

We additionally observe that \ours{}-L and \ours{}-C achieve comparable results overall, with \ours{}-C performing slightly better on average.
This suggests that for many of our test tasks, the loss surface of the bias parameters is close to linear, enabling a single inferred gradient direction to consistently reduce the loss.
Nonetheless, the more pronounced improvement observed on some tasks (\emph{e.g.,} Traffic Sign and ISIC) highlights the potential benefit of modeling non-linear behavior in the loss surface.

Finally, ProtoNet + Bias Tuning performs comparably to ProtoNet + Full Tuning, which shows that updating only the bias parameters does not restrict the flexibility of the few-shot adaptation.
The parameter efficiency of bias tuning is key to the success of our method, enabling significant performance gains with minimal computational overhead.

\subsection{Computation Cost Analysis}
\label{sec:cost_analysis}
In this section, we provide a detailed comparison of the memory and computation time of \ours{} against direct-transfer and fine-tuning baselines.
Since the computation costs of Full-Tuning and Bias-Tuning are about the same, we report the Full-Tuning as ProtoNet Fine-Tuning.

\vspace{-0.3cm}
\paragraph{Memory Efficiency}
Table~\ref{tab:memory_cost} reports the peak memory usage of each method when processing a single 40-way, 5-shot episode with $224 \times 224$ images.
We observe that \ours{} reduces the memory requirement to just $6\%$ of that used by fine-tuning, introducing an overhead of only $60\%$ compared to the vanilla ProtoNet inference.
This substantial gain in memory efficiency stems from the fact that \ours{} does not need to store the entire computation graph—its gradient-free adaptation runs entirely in inference mode.
As a result, \ours{} is particularly appealing for adapting large-scale models in resource-constrained environments, where fine-tuning would be infeasible due to the memory constraint.

\begin{figure}[!t]
    \centering
    \includegraphics[width=\linewidth]{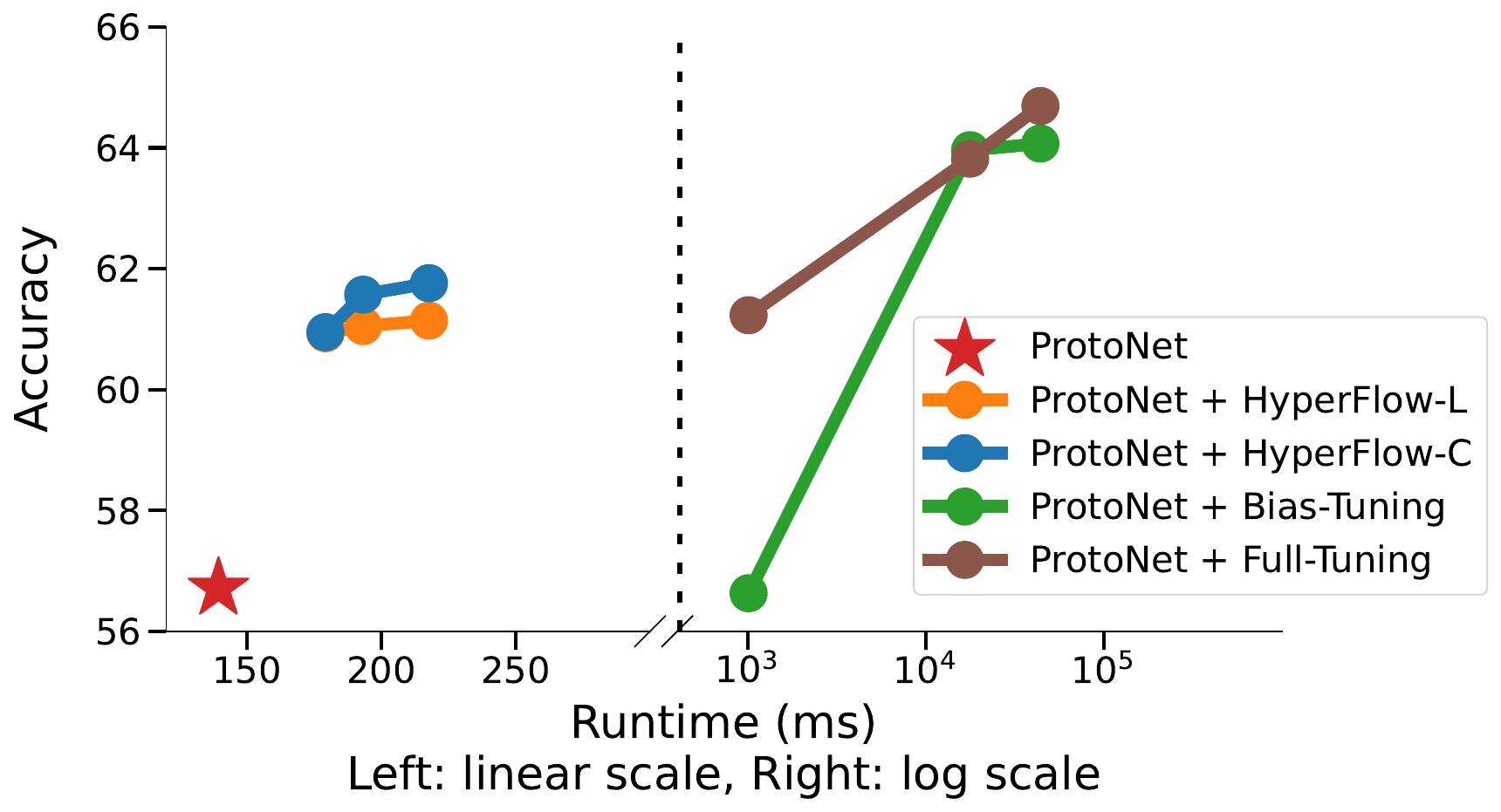}
    \vspace{-0.4cm}
    \caption{Average 5-shot CDFSL performance versus computation cost.
    For \ours{} and the fine-tuning baselines, we plot the total time spent for inference and adaptation with three different number of update steps: 1, 20, and 50 (for Euler solver and gradient descent, respectively).
    \ours{} offers an effective middle ground between direct-transfer and fine-tuning approaches.}
    \label{fig:inference_time}
    \vspace{-0.3cm}
\end{figure}

\vspace{-0.3cm}
\paragraph{Time Efficiency}
Table~\ref{tab:inference_time} shows the average runtime of each method on an NVIDIA A6000 GPU for the same 40-way, 5-shot episode.
By decoupling the adaptation process from the target mode, \ours{} requires merely $0.2\%$ of the computational cost of the fine-tuning approach, translating to only $1.7\times$ increase over the vanilla ProtoNet inference time.
As illustrated in Figure~\ref{fig:inference_time}, \ours{} thus occupies the middle ground between direct-transfer and fine-tuning in both performance and computation cost.
It is noticeable that \ours{} with 50-steps is even $10\times$ faster than the single-step fine-tuning while achieving better performance.
These results suggest that \ours{} can be highly advantageous in applications that demand frequent or real-time adaptation.

\begin{table}[!t]
    \caption{A comparison of memory requirements. We measure the peak memory on a 40-way, 5-shot episode with $224 \times 224$ images.}
    \vspace{-0.1cm}
    \label{tab:memory_cost}
    \renewcommand{\arraystretch}{1.2}
    \renewcommand{\aboverulesep}{0pt}
    \renewcommand{\belowrulesep}{0pt}
    \setlength\tabcolsep{4pt}
    \small
    \centering
    \begin{tabular}{c|c}
         \toprule
         Model Variant & Memory (MB)  \\
         \midrule
         ProtoNet Inference &  900.28 \\
         \midrule
         \ours{} ODE-Solving &  1447.77 \\
         ProtoNet Fine-Tuning &  23118.26 \\
         \bottomrule
    \end{tabular}
    \vspace{-0.1cm}
\end{table}
\begin{table}[!t]
    \caption{A comparison of computation time. We measure the average runtime on a 40-way, 5-shot episode with $224 \times 224$ images.
    We use a NVIDIA A6000 GPU to report the average over 5 runs.}
    \vspace{-0.1cm}
    \label{tab:inference_time}
    \renewcommand{\arraystretch}{1.2}
    \renewcommand{\aboverulesep}{0pt}
    \renewcommand{\belowrulesep}{0pt}
    \setlength\tabcolsep{4pt}
    \small
    \centering
    \begin{tabular}{c|c}
         \toprule
         Model Variant & Time (ms)  \\
         \midrule
         ProtoNet Inference &  139.46 \\
         \midrule
         \ours{} ODE-Solving (50 steps) &  80.10 \\
         ProtoNet Fine-Tuning (1 step) & 874.40 \\
         ProtoNet Fine-Tuning (50 steps) &  43766.67 \\
         \bottomrule
    \end{tabular}
    \vspace{-0.2cm}
\end{table}

\vspace{-0.3cm}
\paragraph{\ours{}-L vs. \ours{}-C}
From Figure~\ref{fig:inference_time}, we see that the performance of \ours{}-L steadily improves with additional inference steps, whereas \ours{}-C gains more pronounced benefits from extended adaptation.
This pattern aligns with the linear flow objective, which encourages a \emph{straight} trajectory through the ODE solution, reducing the gap between single-step and multi-step adaptation.
In highly resource-constrained scenarios that permit only a few inference steps of the drift network, \ours{}-L can be attractive, as it does not require storing full parameter trajectories and thus has a much lower storage overhead than \ours{}-C.
On the other hand, \ours{}-C offers better flexibility by trading off additional computation for higher accuracy, thereby underscoring \ours{}’s effectiveness as a middle ground between direct-transfer and full fine-tuning.

\subsection{Ablation Study}
\label{sec:ablation_study}

\paragraph{Effect of ODE-based Inference}
To investigate the impact of the ODE-based adaptation mechanism, we compare it against a non-ODE variant (\textbf{HyperNet}).
Both methods share the same training trajectories and architecture, but HyperNet directly predicts the final parameters $\theta_T$ rather than iteratively updating them.
It can also be seen as a simplified version of \ours{}-L where it only predicts the drift at the initial point of the trajectories.
Table~\ref{tab:ablation_ode_table} summarizes their performance on the Meta-Dataset and CDFSL benchmarks, showing that the ODE-based approaches consistently outperform HyperNet.
We attribute this improvement to the iterative nature of \ours{}, which allows trading additional computation for more accurate parameter updates.

\vspace{-0.3cm}
\paragraph{Effect of Trajectory Diversity}
To further examine where \ours{}’s generalization ability originates, we conduct ablation studies that vary the size and diversity of the training trajectories.
In Table~\ref{tab:ablation_domain_table} and Table~\ref{tab:ablation_task_table}, we show results when the conditional drift network is trained on different numbers of domains and tasks, respectively.
As expected, performance increases consistently with larger and more diverse training trajectories, emphasizing that a broad base dataset is crucial for robust generalization to new domains and tasks.
Additionally, Table~\ref{tab:ablation_seed_table} explores the effect of varying the number of trajectories per task with different initializations while keeping the total number of tasks fixed.
We find that using only a single trajectory per task degrades the performance, indicating that broader exploration of the parameter space is vital for the drift network’s generalization.

\begin{table}[!t]
\caption{Comparison with a non-ODE variant of \ours{} in Meta-Dataset and CDFSL benchmarks.}
\label{tab:ablation_ode_table}
\vspace{-0.4cm}
\begin{center}
    \renewcommand{\arraystretch}{1.2}
    \renewcommand{\aboverulesep}{0pt}
    \renewcommand{\belowrulesep}{0pt}
    \setlength\tabcolsep{4pt}
    \footnotesize
    \begin{tabular}{c|ccc|ccc}
        \toprule
        \multirow{2}{*}{Model Variants} &
        \multicolumn{3}{c|}{Meta-Dataset average} &
        \multicolumn{3}{c}{CDFSL average } \\
        \cmidrule{2-7} & Total & OOD & ID & 5w5s & 5w20s & 5w50s \\
        \midrule
        HyperNet & 78.64 & 58.83 & 83.59 & 61.07 & 66.30 & 67.96 \\
        \ours{}-L & 79.10 & 58.97 & 84.13 & 61.14 & 66.34 & 68.10 \\
        \ours{}-C & 79.26 & 61.09 & 83.88 & 61.76 & 67.22 & 69.01 \\

        \bottomrule
    \end{tabular}
\end{center}
\vspace{-0.1cm}
\end{table}
\begin{table}[!t]
\caption{Ablation study on the number of domains used to train the conditional drift network.}
\label{tab:ablation_domain_table}
\vspace{-0.4cm}
\begin{center}
    \renewcommand{\arraystretch}{1.2}
    \renewcommand{\aboverulesep}{0pt}
    \renewcommand{\belowrulesep}{0pt}
    \setlength\tabcolsep{4pt}
    \footnotesize
    \begin{tabular}{c|ccc|ccc}
        \toprule
        \multirow{2}{*}{\# of Domains} &
        \multicolumn{3}{c|}{Meta-Dataset average} &
        \multicolumn{3}{c}{CDFSL average } \\
        \cmidrule{2-7} & Total & OOD & ID & 5w5s & 5w20s & 5w50s \\
        \midrule
        1 & 77.80 & 55.52 & 83.37 & 60.68 & 65.72 & 68.24 \\
        4 & 77.84 & 55.81 & 83.34 & 61.69 & 66.82 & 68.53 \\
        8 & 79.26 & 61.09 & 83.88 & 61.76 & 67.22 & 68.54 \\

        \bottomrule
    \end{tabular}
\end{center}
\vspace{-0.1cm}
\end{table}
\begin{table}[!t]
\caption{Ablation study on the number of tasks used to train the conditional drift network.}
\label{tab:ablation_task_table}
\vspace{-0.4cm}
\begin{center}
    \renewcommand{\arraystretch}{1.2}
    \renewcommand{\aboverulesep}{0pt}
    \renewcommand{\belowrulesep}{0pt}
    \setlength\tabcolsep{4pt}
    \footnotesize
    \begin{tabular}{c|ccc|ccc}
        \toprule
        \multirow{2}{*}{\# of tasks} &
        \multicolumn{3}{c|}{Meta-Dataset average} &
        \multicolumn{3}{c}{CDFSL average } \\
        \cmidrule{2-7} & Total & OOD & ID & 5w5s & 5w20s & 5w50s \\
        \midrule
        100 & 77.71 & 55.13 & 83.35 & 60.42 & 65.59 & 67.68 \\
        500 & 79.26 & 61.09 & 83.88 & 61.76 & 67.22 & 68.54 \\

        \bottomrule
    \end{tabular}
\end{center}
\vspace{-0.1cm}
\end{table}
\begin{table}[!t]
\caption{Ablation study on the number of initializations of fine-tuning trajectories used to train the conditional drift network.}
\label{tab:ablation_seed_table}
\vspace{-0.4cm}
\begin{center}
    \renewcommand{\arraystretch}{1.2}
    \renewcommand{\aboverulesep}{0pt}
    \renewcommand{\belowrulesep}{0pt}
    \setlength\tabcolsep{4pt}
    \footnotesize
    \begin{tabular}{c|ccc|ccc}
        \toprule
        \multirow{2}{*}{\# of initializations} &
        \multicolumn{3}{c|}{Meta-Dataset average} &
        \multicolumn{3}{c}{CDFSL average } \\
        \cmidrule{2-7} & Total & OOD & ID & 5w5s & 5w20s & 5w50s \\
        \midrule
        1 & 78.25 & 56.35 & 83.72 & 59.32 & 63.78 & 66.82 \\
        10 & 79.26 & 61.09 & 83.88 & 61.76 & 67.22 & 68.54 \\

        \bottomrule
    \end{tabular}
\end{center}
\vspace{-0.2cm}
\end{table}
\section{Conclusion}
\label{sec:conclusion}
We introduced \ours{} as an alternative to the two predominant few-shot learning paradigms: direct transfer and fine-tuning.
By formulating the gradient descent procedure of fine-tuning as an ordinary differential equation, we designed a conditional drift network that predicts the velocity field of the parameters given a few support examples.
We focused updates solely on the bias parameters, enabling efficient training and inference on the parameter space.
We also proposed two ODE objectives that smoothly interpolate the discrete trajectories of gradient-descent simulations on the base dataset.
Through extensive cross-domain few-shot classification experiments, we demonstrated that \ours{} significantly improves out-of-domain performance while adding minimal computational overhead to the vanilla target model.
As a result, \ours{} effectively occupies the middle ground between direct transfer and full fine-tuning in terms of both accuracy and computational cost.

\clearpage
{
    \small
    \bibliographystyle{ieeenat_fullname}
    \bibliography{main}

\begin{thebibliography}{55}
\providecommand{\natexlab}[1]{#1}
\providecommand{\url}[1]{\texttt{#1}}
\expandafter\ifx\csname urlstyle\endcsname\relax
  \providecommand{\doi}[1]{doi: #1}\else
  \providecommand{\doi}{doi: \begingroup \urlstyle{rm}\Url}\fi

\bibitem[Andrychowicz et~al.(2016)Andrychowicz, Denil, Gomez, Hoffman, Pfau, Schaul, Shillingford, and de~Freitas]{andrychowicz2016learning}
Marcin Andrychowicz, Misha Denil, Sergio Gomez, Matthew~W Hoffman, David Pfau, Tom Schaul, Brendan Shillingford, and Nando de Freitas.
\newblock Learning to learn by gradient descent by gradient descent.
\newblock In \emph{Advances in Neural Information Processing Systems (NeurIPS)}, pages 3988--3996, 2016.

\bibitem[Arg{\"u}eso et~al.(2020)Arg{\"u}eso, Picon, Irusta, Medela, San-Emeterio, Bereciartua, and Alvarez-Gila]{argueso2020few}
David Arg{\"u}eso, Artzai Picon, Unai Irusta, Alfonso Medela, Miguel~G San-Emeterio, Arantza Bereciartua, and Aitor Alvarez-Gila.
\newblock Few-shot learning approach for plant disease classification using images taken in the field.
\newblock \emph{Computers and Electronics in Agriculture}, 175:\penalty0 105542, 2020.

\bibitem[Bateni et~al.(2020)Bateni, Goyal, Masrani, Wood, and Sigal]{bateni2020improved}
Peyman Bateni, Raghav Goyal, Vaden Masrani, Frank Wood, and Leonid Sigal.
\newblock Improved few-shot visual classification.
\newblock In \emph{Proceedings of the IEEE/CVF conference on computer vision and pattern recognition}, pages 14493--14502, 2020.

\bibitem[Bommasani et~al.(2021)Bommasani, Hudson, Adeli, Altman, Arora, von Arx, Bernstein, Bohg, Bosselut, Brunskill, et~al.]{bommasani2021opportunities}
Rishi Bommasani, Drew~A Hudson, Ehsan Adeli, Russ Altman, Simran Arora, Sydney von Arx, Michael~S Bernstein, Jeannette Bohg, Antoine Bosselut, Emma Brunskill, et~al.
\newblock On the opportunities and risks of foundation models.
\newblock \emph{arXiv preprint arXiv:2108.07258}, 2021.

\bibitem[Bu et~al.(2021)Bu, Xu, and Chen]{bu2021dynamical}
Zhiqi Bu, Shiyun Xu, and Kan Chen.
\newblock A dynamical view on optimization algorithms of overparameterized neural networks.
\newblock In \emph{International conference on artificial intelligence and statistics}, pages 3187--3195. PMLR, 2021.

\bibitem[Caron et~al.(2021)Caron, Touvron, Misra, J{\'e}gou, Mairal, Bojanowski, and Joulin]{caron2021emerging}
Mathilde Caron, Hugo Touvron, Ishan Misra, Herv{\'e} J{\'e}gou, Julien Mairal, Piotr Bojanowski, and Armand Joulin.
\newblock Emerging properties in self-supervised vision transformers.
\newblock In \emph{Proceedings of the IEEE/CVF international conference on computer vision}, pages 9650--9660, 2021.

\bibitem[Catmull and Rom(1974)]{catmull1974class}
Edwin Catmull and Raphael Rom.
\newblock A class of local interpolating splines.
\newblock In \emph{Computer aided geometric design}, pages 317--326. Elsevier, 1974.

\bibitem[Chen et~al.(2018)Chen, Rubanova, Bettencourt, and Duvenaud]{chen2018neural}
Ricky~TQ Chen, Yulia Rubanova, Jesse Bettencourt, and David~K Duvenaud.
\newblock Neural ordinary differential equations.
\newblock \emph{Advances in neural information processing systems}, 31, 2018.

\bibitem[Chen et~al.(2023)Chen, Duan, Wang, He, Lu, Dai, and Qiao]{chen2023vision}
Zhe Chen, Yuchen Duan, Wenhai Wang, Junjun He, Tong Lu, Jifeng Dai, and Yu Qiao.
\newblock Vision transformer adapter for dense predictions.
\newblock In \emph{The Eleventh International Conference on Learning Representations}, 2023.

\bibitem[De~Boor(1978)]{de1978practical}
Carl De~Boor.
\newblock \emph{A practical guide to splines}.
\newblock springer New York, 1978.

\bibitem[Deng et~al.(2009)Deng, Dong, Socher, Li, Li, and Fei-Fei]{deng2009imagenet}
Jia Deng, Wei Dong, Richard Socher, Li-Jia Li, Kai Li, and Li Fei-Fei.
\newblock Imagenet: A large-scale hierarchical image database.
\newblock In \emph{2009 IEEE conference on computer vision and pattern recognition}, pages 248--255. Ieee, 2009.

\bibitem[Dosovitskiy et~al.(2021)Dosovitskiy, Beyer, Kolesnikov, Weissenborn, Zhai, Unterthiner, Dehghani, Minderer, Heigold, Gelly, et~al.]{dosovitskiy2021image}
Alexey Dosovitskiy, Lucas Beyer, Alexander Kolesnikov, Dirk Weissenborn, Xiaohua Zhai, Thomas Unterthiner, Mostafa Dehghani, Matthias Minderer, Georg Heigold, Sylvain Gelly, et~al.
\newblock An image is worth 16x16 words: Transformers for image recognition at scale.
\newblock In \emph{International Conference on Learning Representations}, 2021.

\bibitem[Du et~al.(2023)Du, Xiao, Liao, and Snoek]{du2023protodiff}
Yingjun Du, Zehao Xiao, Shengcai Liao, and Cees Snoek.
\newblock Protodiff: Learning to learn prototypical networks by task-guided diffusion.
\newblock \emph{Advances in Neural Information Processing Systems}, 36:\penalty0 46304--46322, 2023.

\bibitem[Elkabetz and Cohen(2021)]{elkabetz2021continuous}
Omer Elkabetz and Nadav Cohen.
\newblock Continuous vs. discrete optimization of deep neural networks.
\newblock \emph{Advances in Neural Information Processing Systems}, 34:\penalty0 4947--4960, 2021.

\bibitem[Fei-Fei et~al.(2006)Fei-Fei, Fergus, and Perona]{fei2006one}
Li Fei-Fei, Robert Fergus, and Pietro Perona.
\newblock One-shot learning of object categories.
\newblock \emph{IEEE transactions on pattern analysis and machine intelligence}, 28\penalty0 (4):\penalty0 594--611, 2006.

\bibitem[Finn et~al.(2017)Finn, Abbeel, and Levine]{finn2017model}
Chelsea Finn, Pieter Abbeel, and Sergey Levine.
\newblock Model-agnostic meta-learning for fast adaptation of deep networks.
\newblock In \emph{International conference on machine learning}, pages 1126--1135. PMLR, 2017.

\bibitem[Guo et~al.(2020)Guo, Codella, Karlinsky, Codella, Smith, Saenko, Rosing, and Feris]{guo2020broader}
Yunhui Guo, Noel~C Codella, Leonid Karlinsky, James~V Codella, John~R Smith, Kate Saenko, Tajana Rosing, and Rogerio Feris.
\newblock A broader study of cross-domain few-shot learning.
\newblock In \emph{Computer vision--ECCV 2020: 16th European conference, glasgow, UK, August 23--28, 2020, proceedings, part XXVII 16}, pages 124--141. Springer, 2020.

\bibitem[Ha et~al.(2017)Ha, Dai, and Le]{ha2017hypernetworks}
David Ha, Andrew~M Dai, and Quoc~V Le.
\newblock Hypernetworks.
\newblock In \emph{International Conference on Learning Representations}, 2017.

\bibitem[He et~al.(2016)He, Zhang, Ren, and Sun]{he2016deep}
Kaiming He, Xiangyu Zhang, Shaoqing Ren, and Jian Sun.
\newblock Deep residual learning for image recognition.
\newblock In \emph{Proceedings of the IEEE conference on computer vision and pattern recognition}, pages 770--778, 2016.

\bibitem[Hermite(1863)]{hermite1863remarque}
C Hermite.
\newblock Remarque sur le d{\'e}veloppement de cosamx.
\newblock \emph{Comptes Rendus de l'Acad{\'e}mie des Sciences}, 57:\penalty0 613--618, 1863.

\bibitem[Ho et~al.(2020)Ho, Jain, and Abbeel]{ho2020denoising}
Jonathan Ho, Ajay Jain, and Pieter Abbeel.
\newblock Denoising diffusion probabilistic models.
\newblock \emph{Advances in neural information processing systems}, 33:\penalty0 6840--6851, 2020.

\bibitem[Hu et~al.(2022{\natexlab{a}})Hu, Wallis, Allen-Zhu, Li, Wang, Wang, Chen, et~al.]{hu2022lora}
Edward~J Hu, Phillip Wallis, Zeyuan Allen-Zhu, Yuanzhi Li, Shean Wang, Lu Wang, Weizhu Chen, et~al.
\newblock Lora: Low-rank adaptation of large language models.
\newblock In \emph{International Conference on Learning Representations}, 2022{\natexlab{a}}.

\bibitem[Hu et~al.(2022{\natexlab{b}})Hu, Li, St{\"u}hmer, Kim, and Hospedales]{hu2022pushing}
Shell~Xu Hu, Da Li, Jan St{\"u}hmer, Minyoung Kim, and Timothy~M Hospedales.
\newblock Pushing the limits of simple pipelines for few-shot learning: External data and fine-tuning make a difference.
\newblock In \emph{Proceedings of the IEEE/CVF Conference on Computer Vision and Pattern Recognition}, pages 9068--9077, 2022{\natexlab{b}}.

\bibitem[Islam et~al.(2021)Islam, Chen, Panda, Karlinsky, Feris, and Radke]{islam2021dynamic}
Ashraful Islam, Chun-Fu~Richard Chen, Rameswar Panda, Leonid Karlinsky, Rogerio Feris, and Richard~J Radke.
\newblock Dynamic distillation network for cross-domain few-shot recognition with unlabeled data.
\newblock \emph{Advances in Neural Information Processing Systems}, 34:\penalty0 3584--3595, 2021.

\bibitem[Jin et~al.(2024)Jin, Wang, Tang, Zhao, Zhou, Tang, and You]{jin2024conditional}
Xiaolong Jin, Kai Wang, Dongwen Tang, Wangbo Zhao, Yukun Zhou, Junshu Tang, and Yang You.
\newblock Conditional lora parameter generation.
\newblock \emph{CoRR}, 2024.

\bibitem[Kim et~al.(2023)Kim, Kim, Cho, Luo, and Hong]{kim2023universal}
Donggyun Kim, Jinwoo Kim, Seongwoong Cho, Chong Luo, and Seunghoon Hong.
\newblock Universal few-shot learning of dense prediction tasks with visual token matching.
\newblock In \emph{The Eleventh International Conference on Learning Representations}, 2023.

\bibitem[Kim et~al.(2024)Kim, Cho, Kim, Luo, and Hong]{kim2024chameleon}
Donggyun Kim, Seongwoong Cho, Semin Kim, Chong Luo, and Seunghoon Hong.
\newblock Chameleon: A data-efficient generalist for dense visual prediction in the wild.
\newblock In \emph{European Conference on Computer Vision}, pages 422--441. Springer, 2024.

\bibitem[Kingma and Ba(2015)]{kingma2015adam}
Diederik~P. Kingma and Jimmy Ba.
\newblock Adam: A method for stochastic optimization.
\newblock In \emph{ICLR (Poster)}, 2015.

\bibitem[Knyazev et~al.(2021)Knyazev, Drozdzal, Taylor, and Romero~Soriano]{knyazev2021parameter}
Boris Knyazev, Michal Drozdzal, Graham~W Taylor, and Adriana Romero~Soriano.
\newblock Parameter prediction for unseen deep architectures.
\newblock \emph{Advances in Neural Information Processing Systems}, 34:\penalty0 29433--29448, 2021.

\bibitem[Kotia et~al.(2020)Kotia, Kotwal, Bharti, and Mangrulkar]{kotia2020few}
Jai Kotia, Adit Kotwal, Rishika Bharti, and Ramchandra Mangrulkar.
\newblock Few shot learning for medical imaging.
\newblock In \emph{Machine learning algorithms for industrial applications}, pages 107--132. Springer, 2020.

\bibitem[Lake et~al.(2015)Lake, Salakhutdinov, and Tenenbaum]{lake2015human}
Brenden~M Lake, Ruslan Salakhutdinov, and Joshua~B Tenenbaum.
\newblock Human-level concept learning through probabilistic program induction.
\newblock \emph{Science}, 350\penalty0 (6266):\penalty0 1332--1338, 2015.

\bibitem[Liang et~al.(2024)Liang, Xu, Hu, Jiang, Huang, and Xu]{liang2024make}
Yongyuan Liang, Tingqiang Xu, Kaizhe Hu, Guangqi Jiang, Furong Huang, and Huazhe Xu.
\newblock Make-an-agent: A generalizable policy network generator with behavior-prompted diffusion.
\newblock In \emph{The Thirty-eighth Annual Conference on Neural Information Processing Systems}, 2024.

\bibitem[Lipman et~al.(2023)Lipman, Chen, Ben-Hamu, Nickel, and Le]{lipman2023flow}
Yaron Lipman, Ricky~TQ Chen, Heli Ben-Hamu, Maximilian Nickel, and Matthew Le.
\newblock Flow matching for generative modeling.
\newblock In \emph{The Eleventh International Conference on Learning Representations}, 2023.

\bibitem[Liu et~al.(2024)Liu, Zhang, Dai, Zhang, Cai, Zhou, and Chen]{liu2024few}
Fan Liu, Tianshu Zhang, Wenwen Dai, Chuanyi Zhang, Wenwen Cai, Xiaocong Zhou, and Delong Chen.
\newblock Few-shot adaptation of multi-modal foundation models: A survey.
\newblock \emph{Artificial Intelligence Review}, 57\penalty0 (10):\penalty0 268, 2024.

\bibitem[Liu et~al.(2023)Liu, Gong, et~al.]{liu2023flow}
Xingchao Liu, Chengyue Gong, et~al.
\newblock Flow straight and fast: Learning to generate and transfer data with rectified flow.
\newblock In \emph{The Eleventh International Conference on Learning Representations}, 2023.

\bibitem[Madan et~al.(2025)Madan, Peri, Kong, and Ramanan]{madan2025revisiting}
Anish Madan, Neehar Peri, Shu Kong, and Deva Ramanan.
\newblock Revisiting few-shot object detection with vision-language models.
\newblock \emph{Advances in Neural Information Processing Systems}, 37:\penalty0 19547--19560, 2025.

\bibitem[Merkulov and Oseledets(2020)]{merkulov2020stochastic}
Daniil Merkulov and Ivan Oseledets.
\newblock Stochastic gradient algorithms from ode splitting perspective.
\newblock In \emph{ICLR 2020 Workshop on Integration of Deep Neural Models and Differential Equations}, 2020.

\bibitem[Oh et~al.(2022)Oh, Kim, Ho, Kim, Song, and Yun]{oh2022understanding}
Jaehoon Oh, Sungnyun Kim, Namgyu Ho, Jin-Hwa Kim, Hwanjun Song, and Se-Young Yun.
\newblock Understanding cross-domain few-shot learning based on domain similarity and few-shot difficulty.
\newblock \emph{Advances in Neural Information Processing Systems}, 35:\penalty0 2622--2636, 2022.

\bibitem[Peebles et~al.(2022)Peebles, Radosavovic, Brooks, Efros, and Malik]{peebles2022learning}
William~S Peebles, Ilija Radosavovic, Tim Brooks, Alexei~A Efros, and Jitendra Malik.
\newblock Learning to learn with generative models of neural network checkpoints.
\newblock \emph{CoRR}, 2022.

\bibitem[Phoo and Hariharan(2021)]{phoo2021self}
Cheng~Perng Phoo and Bharath Hariharan.
\newblock Self-training for few-shot transfer across extreme task differences.
\newblock In \emph{International Conference on Learning Representations}, 2021.

\bibitem[Ravi and Larochelle(2017)]{ravi2017optimization}
Sachin Ravi and Hugo Larochelle.
\newblock Optimization as a model for few-shot learning.
\newblock In \emph{International conference on learning representations}, 2017.

\bibitem[Rebuffi et~al.(2017)Rebuffi, Bilen, and Vedaldi]{rebuffi2017learning}
Sylvestre-Alvise Rebuffi, Hakan Bilen, and Andrea Vedaldi.
\newblock Learning multiple visual domains with residual adapters.
\newblock \emph{Advances in neural information processing systems}, 30, 2017.

\bibitem[Rombach et~al.(2022)Rombach, Blattmann, Lorenz, Esser, and Ommer]{rombach2022high}
Robin Rombach, Andreas Blattmann, Dominik Lorenz, Patrick Esser, and Bj{\"o}rn Ommer.
\newblock High-resolution image synthesis with latent diffusion models.
\newblock In \emph{Proceedings of the IEEE/CVF conference on computer vision and pattern recognition}, pages 10684--10695, 2022.

\bibitem[Santambrogio(2017)]{santambrogio2017euclidean}
Filippo Santambrogio.
\newblock $\{$Euclidean, metric, and Wasserstein$\}$ gradient flows: an overview.
\newblock \emph{Bulletin of Mathematical Sciences}, 7:\penalty0 87--154, 2017.

\bibitem[Sch{\"u}rholt et~al.(2022)Sch{\"u}rholt, Knyazev, Gir{\'o}-i Nieto, and Borth]{schurholt2022hyper}
Konstantin Sch{\"u}rholt, Boris Knyazev, Xavier Gir{\'o}-i Nieto, and Damian Borth.
\newblock Hyper-representations as generative models: Sampling unseen neural network weights.
\newblock \emph{Advances in Neural Information Processing Systems}, 35:\penalty0 27906--27920, 2022.

\bibitem[Sch{\"u}rholt et~al.(2024)Sch{\"u}rholt, Mahoney, and Borth]{schurhol2024ttowards}
Konstantin Sch{\"u}rholt, Michael~W Mahoney, and Damian Borth.
\newblock Towards scalable and versatile weight space learning.
\newblock In \emph{Forty-first International Conference on Machine Learning}, 2024.

\bibitem[Snell et~al.(2017)Snell, Swersky, and Zemel]{snell2017prototypical}
Jake Snell, Kevin Swersky, and Richard Zemel.
\newblock Prototypical networks for few-shot learning.
\newblock \emph{Advances in neural information processing systems}, 30, 2017.

\bibitem[Sun et~al.(2021)Sun, Wang, Wang, Li, Li, and Fu]{sun2021research}
Xian Sun, Bing Wang, Zhirui Wang, Hao Li, Hengchao Li, and Kun Fu.
\newblock Research progress on few-shot learning for remote sensing image interpretation.
\newblock \emph{IEEE Journal of Selected Topics in Applied Earth Observations and Remote Sensing}, 14:\penalty0 2387--2402, 2021.

\bibitem[Triantafillou et~al.(2020)Triantafillou, Zhu, Dumoulin, Lamblin, Evci, Xu, Goroshin, Gelada, Swersky, Manzagol, et~al.]{triantafillou2020meta}
Eleni Triantafillou, Tyler Zhu, Vincent Dumoulin, Pascal Lamblin, Utku Evci, Kelvin Xu, Ross Goroshin, Carles Gelada, Kevin Swersky, Pierre-Antoine Manzagol, et~al.
\newblock Meta-dataset: A dataset of datasets for learning to learn from few examples.
\newblock In \emph{International Conference on Learning Representations}, 2020.

\bibitem[Vaswani et~al.(2017)Vaswani, Shazeer, Parmar, Uszkoreit, Jones, Gomez, Kaiser, and Polosukhin]{vaswani2017attention}
Ashish Vaswani, Noam Shazeer, Niki Parmar, Jakob Uszkoreit, Llion Jones, Aidan~N Gomez, {\L}ukasz Kaiser, and Illia Polosukhin.
\newblock Attention is all you need.
\newblock \emph{Advances in neural information processing systems}, 30, 2017.

\bibitem[Vinyals et~al.(2016)Vinyals, Blundell, Lillicrap, Wierstra, et~al.]{vinyals2016matching}
Oriol Vinyals, Charles Blundell, Timothy Lillicrap, Daan Wierstra, et~al.
\newblock Matching networks for one shot learning.
\newblock \emph{Advances in neural information processing systems}, 29, 2016.

\bibitem[Xu et~al.(2024)Xu, Shi, Wei, Mu, Li, and Liang]{xu2024towards}
Zhuoyan Xu, Zhenmei Shi, Junyi Wei, Fangzhou Mu, Yin Li, and Yingyu Liang.
\newblock Towards few-shot adaptation of foundation models via multitask finetuning.
\newblock In \emph{The Twelfth International Conference on Learning Representations}, 2024.

\bibitem[Zaken et~al.(2022)Zaken, Goldberg, and Ravfogel]{zaken2022bitfit}
Elad~Ben Zaken, Yoav Goldberg, and Shauli Ravfogel.
\newblock Bitfit: Simple parameter-efficient fine-tuning for transformer-based masked language-models.
\newblock In \emph{Proceedings of the 60th Annual Meeting of the Association for Computational Linguistics (Volume 2: Short Papers)}, pages 1--9, 2022.

\bibitem[Zhang et~al.(2022)Zhang, Li, Feng, Ye, and Ye]{zhang2022metanode}
Baoquan Zhang, Xutao Li, Shanshan Feng, Yunming Ye, and Rui Ye.
\newblock Metanode: Prototype optimization as a neural ode for few-shot learning.
\newblock In \emph{Proceedings of the AAAI Conference on Artificial Intelligence}, pages 9014--9021, 2022.

\bibitem[Zhang et~al.(2024)Zhang, Luo, Yu, Li, Lin, Ye, and Zhang]{zhang2024metadiff}
Baoquan Zhang, Chuyao Luo, Demin Yu, Xutao Li, Huiwei Lin, Yunming Ye, and Bowen Zhang.
\newblock Metadiff: Meta-learning with conditional diffusion for few-shot learning.
\newblock In \emph{Proceedings of the AAAI conference on artificial intelligence}, pages 16687--16695, 2024.

\end{thebibliography}
}

\end{document}